# COMPARISON OF MACHINE LEARNING FOR SENTIMENT ANALYSIS IN DETECTING ANXIETY BASED ON SOCIAL MEDIA DATA

Shoffan Saifullah[1,a], Yuli Fauziah[1,b], Agus Sasmito Aribowo[1,2,c]

[1]Department of Informatics Engineering, Universitas Pembangunan Nasional Veteran Yogyakarta, Indonesia

[2]Faculty of Information & Communication Technology, Universiti Teknikal Malaysia Melaka, Malaysia

[a]shoffans@upnyk.ac.id; [b]yuli.fauziah@upnyk.ac.id; [c]sasmito.skom@upnyk.ac.id

## Abstract

All groups of people felt the impact of the COVID-19 pandemic. This situation triggers anxiety, which is bad for everyone. The government's role is very influential in solving these problems with its work program. It also has many pros and cons that cause public anxiety. For that, it is necessary to detect anxiety to improve government programs that can increase public expectations. This study applies machine learning to detecting anxiety based on social media comments regarding government programs to deal with this pandemic. This concept will adopt a sentiment analysis in detecting anxiety based on positive and negative comments from netizens. The machine learning methods implemented include K-NN, Bernoulli, Decision Tree Classifier, Support Vector Classifier, Random Forest, and XG-boost. The data sample used is the result of crawling YouTube comments. The data used amounted to 4862 comments consisting of negative and positive data with 3211 and 1651. Negative data identify anxiety, while positive data identifies hope (not anxious). Machine learning is processed based on feature extraction of count-vectorization and TF-IDF. The results showed that the sentiment data amounted to 3889 and 973 in testing, and training with the greatest accuracy was the random forest with feature extraction of vectorization count and TF-IDF of 84.99% and 82.63%, respectively. The best precision test is K-NN, while the best recall is XG-Boost. Thus, Random Forest is the best accurate to detect someone's anxiety based-on data from social media.

**Keywords:** Anxiety Detection, Machine Learning, Sentiment Analysis, Social Media Data, Text Feature Extraction.

## I. INTRODUCTION

Coronavirus disease 2019 (COVID-19) took the world by storm and became a rapidly expanding pandemic [44]. This virus was initially submitted and identified from the 2019-novel coronavirus (2019-nCoV) report on [45]. This virus spread rapidly throughout the world, originating from Wuhan, China [48]. This pandemic also made Indonesia very fast in early March 2020 [6, 33, 39]. This case creates anxiety and anxiety for everyone, including the community and government in Indonesia. The government plays a significant role in handling pandemics and seeks to neutralize and stop pandemics with various programs [35]. The programs launched are the supply of medical devices [23], provision of assistance [38], and free electricity [34].

These government programs have pros and cons, which can be seen in various news and social media. Comments and feedback on social media are highly visible and indicate whether there is anxiety or not. Everyone can make comments and feedback from all walks of life. News in the Covid-19 pandemic, such as details of the spread, death rates, and government programs for handling it, has a massive impact on the community's psyche, such as anxiety and panic [30]. Thus, this conditions make the panic epidemic on social media and spread more than the COVID-19 epidemic itself [1].

Anxiety can be detected using the concept of sentiment analysis [2, 27, 46], especially in text processing. The study explains that sentiment analysis is used to detect anxiety based on one's social media, both from sharing information and feedback (comments) [25]. The concept of text mining in identifying sentiment has not been explained. Besides, related studies explain that anxiety can be detected based on social media due to the COVID-19 pandemic [17, 24]. A psychologist can analyze anxiety detection. However, along with the development of technology and the concept of text mining, anxiety detection can be done quickly and precisely based on training data using machine learning. This concept process used an analysis of a person's sentiment in responding to the pandemic and government programs in dealing with the COVID-19 pandemic. The analysis is based on text-based on YouTube comments. Machine learning methods are K-NN, Bernoulli, Decision Tree Classifier, Support Vector Classifier, Random Forest, and XG-boost. This process used the feature extraction of text data processing, namely Count-vectorization and TF-IDF.

This article contains some of the five main sections. The first part is a literature review describing related studies regarding machine learning in text mining processing and its differences. The second part is a method for all the processes and the steps and methods used in this research. The fourth explains the results and discussion of this research. The last section contains the conclusions of the research based-on result and discussion.

## II. LITERATURE REVIEW

Text mining can detect human emotions based on data from social media. This research has been done, but it is not satisfactory and is still developing. Emotion detection research divides two emotions, namely hate and non-hate. Besides, the distribution of feelings is also divided into three types: strong hate, weak hate, and no hate [40]. In its identification, this study analyzes emotions in the form of detection of fanaticism in the document text. In the detection of fanaticism, the concept of fanaticism is divided into three, namely non-fanaticism, Code Attitude fanaticism, and Code Red Fanaticism. These concepts are classified using Case-Based Reasoning (CBR) and Naïve Bayes methods with an accuracy of up to 77% [3].

Because the accuracy is still low, the research is continued using the machine learning concept, using the Random Forest method. However, in this study, the accuracy rate fell to 72% [19]. The research was published in Arabic language newspapers, including Al-Jazeera, Al-Arabia, Al-Watan, and Al-Qabas. The implementation of machine learning methods in the classification process reaches 96%. The methods used are C4.5, RIPPER, and PART. These results are included in the structured domain [4]. After that, this research was developed using the feature extraction method. The feature extraction methods used include TF-IDF, Support Vector Machine, and Naïve Bayesian. Implementation with the feature extraction method yields an accuracy of 82.1% [5].

Research in sentiment analysis can detect hate speech based on data from Facebook. This research belongs to unstructured sentiment analysis [40]. The application of the Support Vector Machines (SVM) method and a particular Recurrent Neural Network, namely Long Short Term Memory (LSTM), were tested for performance in emotion recognition. Emotions identified by this method are positive and negative emotions. The study resulted in the effectiveness of both approaches in the Italian language domain. Other studies also apply the lexicon method and a combination of sentiment dictionaries and sentiment corpus in the sentiment detection process. This method is a model for detecting sentiment in English. The test results produce an accuracy of 73% [15]. Besides, sentiment analysis detection can be carried out using two methods: a two-step method for hate speech detection and embedding techniques. The process is carried out by conducting training using a binary classifier that can separate hateful and neutral comments [11].

Emotional detection on social media is carried out using many concepts, one of which is detecting public sentiment in online media. This detection is based on the negative sentiment, which is defined as displeasing utterances related to religion, gender, and specific ethnicity [43]. Besides, related research detects sentiment using the two-step method [11]. Likewise, in other studies, sentiment analysis in the detection process is based on individual objects oriented towards race, nationality, and religion [15]. Analysis of sentiment detection is automatically processed using natural language processing [32] using text data on social media, including analytical sentiment detection on Facebook [40].

The detection of one's expectations is also carried out using social media data in the political domain [8]. The data used is data from Twitter. Researches with the theme of detection of mass anxiety and fear have also been carried out in various circumstances, namely during political wars [10, natural earthquake disasters [41], and in Youtube video comments [9]. However, the results of the research conducted have not shown maximum results. The application of machine learning is not an ensemble method, so this study applies the ensemble method such as Random Forest, XG-Boost, and others.

## III. RESEARCH METHODOLOGY

There are four main steps taken in this research: data collection, system analysis and concepts, and system design. Data collection is done by looking for literature studies to get an overview and literacy related to machine learning and text mining following the problems being solved. The concept of literacy studies is used to solve anxiety cases that exist in Indonesia today, mainly based on comments from social media. This social media can identify comments and feedback from users [29]. Besides, data collection was carried out directly using observation to obtain appropriate data. So that the data processed is following the data obtained on social media. The study's data collection was carried out by observing and taking comments from YouTube videos [26]. The data used is YouTube comment data, which is related to government programs related to free electricity. This study uses data in Indonesian. Every comment from someone is identified about the existing emotion based on positive and negative sentiments. Positive sentiment is optimistic hope, while negative sentiment is someone's anxiety.

This research has a concept of technology and computerization. Besides, the machine learning concept is used in the resolution of anxiety detection cases. This research is developing previous research on machine learning detection but only applies the random forest and XG-Boost methods [12]. The development of this research is the application of feature extraction methods with Count-Vectorization and TF-IDF and machine learning methods using K-NN, Bernoulli, Decision Tree Classifier, Support Vector Classifier, Random Forest, and XG-Boost. System development is carried out using system design techniques, namely prototyping. This technique aims to model a system for detecting sentiment based on social media emotions [7, 14]. The prototyping process uses three main stages in text processing: preprocessing, emotion detection based on sentiment analysis, and cross-validation testing. The preprocessing stage is shown in Figure 1. Preprocessing is carried out to obtain precise text data and can be processed in the next stage. This process begins with the process of crawling data from YouTube comments, then preprocessing. The crawling data is temporarily stored in the database for preprocessing so that the data is normal and detected. This study's preprocessing stages include tokenizing, filtering, stemming, tagging, and the emoticon conversion process. The preprocessing results are in the form of clean text data, which is stored and processed further.

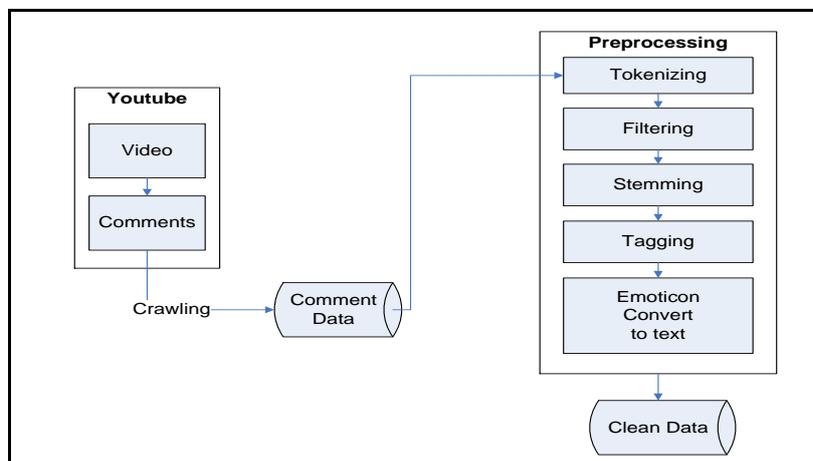

Figure 1. Pre-processing based on YouTube comments regarding the Government Electricity Program

Based on Figure 1, the results are further processed using feature extraction. The feature extraction used includes the Count-Vectorization and TF-IDF (Term Frequency Inverse Document Frequency). Count-Vectorization is a method used to count all the corpus words and make them representations of the document. This method only counts the word frequency. Meanwhile, TF-IDF is a method used to calculate the weight of each word that is commonly used. The method has advantages in several factors, such as efficiency, ease, and accuracy. This method aims to calculate the value of the Term Frequency (TF) and Inverse Document Frequency (IDF) of each token (word) of each document in the corpus. Alternatively, the TF-IDF method serves to find out how often the word appears. It is the first step in the machine learning process stages, according to Figure 2.

Machine learning will process data based on the results of feature extraction. This machine learning process is a concept used to detect sentiments and emotions based on the text whose comments are processed. As stated in the development of previous studies, this study uses five machine learning methods for the detection process. The result will be a training and testing process. The results of the training and test data are calculated for its accuracy using cross-validation. Apart from accuracy, other calculations use precision and recall functions.

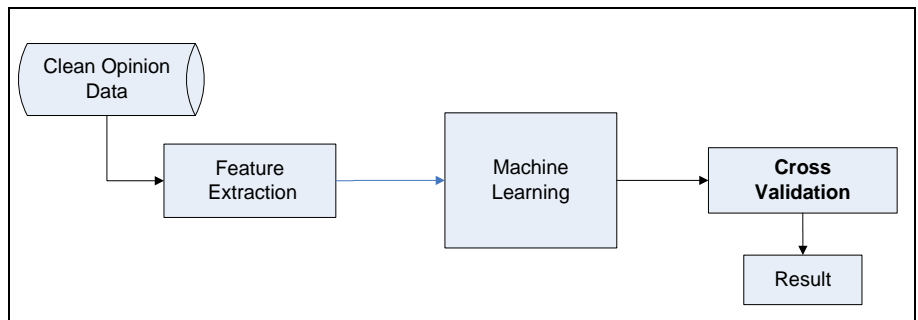

Figure 2. The stages of Machine Learning and the results

Machine learning uses the K-NN (K-Nearest Neighbor) method, Bernoulli, Decision Tree Classifier, Support Vector Classifier, Random Forest, and XG-Boost. KNN is used in the classification process by finding the closest K match based on training data [28]. Furthermore, the closest match label is used in the prediction. In general, the distance used is the euclidean distance in finding the closest match. The second method is Bernoulli. The model is used only to ignore the number of occurrences [21, 42]. Bernoulli's Naïve Bayes classification is a model that specifies that a document is represented by a binary attribute vector that indicates words that appear and do not appear in the document [22].

The decision tree is a method of classification and pattern prediction from data. It describes the relationship of the attribute variable x and target y in the form of a tree. The decision tree has a structure like a flowchart, where the internal node is a test of the attribute variable (not the leaf / outermost), each branch is the test result, and the outer node is the leaf, which is the label [37]. Support Vector Machine (SVM) is a relatively new technique for making predictions, both in classification and regression. Support Vector Machine is included in the supervised learning class. Its implementation must have a training stage using SVM sequential training and followed by the testing phase. The concept of classification with the Support Vector Machine is to find the best hyperplane that functions as a separator of two data classes. The Support Vector Machine can work on high-dimensional datasets using kernel tricks. The Support Vector Machine only uses a few selected data points that contribute (support vector) to form a model used in the classification process [31].

Random Forest (RF) is a classification algorithm that can be used to identify sentiment analysis [36, 20] and emotion analysis [16, 18] in numbers significant data. Random forest is a classification based on combining tree structures with training on the sample data they have. Extreme Gradient Boosting (XG-Boost) is a tree-based classification algorithm development [13]. This algorithm mimics the behavior of Random forest in its tree creation and is also combined with gradient descent / boosting. Gradient Boosting is a machine learning concept in solving regression problems whose classification produces a predictive model in a weak ensemble prediction model [47]. This boosting method can predict the error/residual from the previous model. XG-Boost is the GBM version with advantages in the process; namely, it is efficient and scalable. XG-Boost performs well in performing functions such as regression, classification, and ranking. XG-Boost is also called a tree ensembles algorithm, which collects several classifications and regression trees (CART).

Cross-validation is an iterative validation process in which the dataset is divided into many subsets (sets) of training & validation data. In this study, the method used is the confusion matrix, a method of measuring performance in the classification method. Measurements are processed by sharing the classification results in real terms and processed according to the classification components in Table 1.

Tabel 1.
Confusion Matrix

| Class | Classified Positive | Classified Negative |
|---|---|---|
| Positive | TP (True Positive) | FN (False Negative) |
| Negative | FP (False Positive) | TN (True Negative) |

In this study, the confusion matrix concept used calculates accuracy, precision, and recall. Accuracy is a process that determines the validity of the data based on the calculation between the measurement result and its real value (proximity). Accuracy can be calculated by (1). Apart from accuracy, precision and recall are also implemented in the performance evaluation process in this study. Both methods are calculations that can strengthen the results of the accuracy measurement process. Precision and recall are measurements of data accuracy and are measured by (2), while recall is a measure of the completeness of data and can be measured by (3). Precision is the number of opinion samples "true" label, which means positive sentiment and is divided by the total number of positive sentiment classification samples.

$$accuracy = \frac{TP+TN}{TP+TN+FP+FN} \quad (1)$$

$$precision = \frac{TP}{TP+FP} \quad (1)$$

$$recall = \frac{TP}{TP+FN} \quad (1)$$

## IV. FINDINGS AND DISCUSSION

In this study, there are results and discussions related to the experiments that have been carried out. The details obtained in this experiment include explaining the dataset, the results, and discussions related to the anxiety detection process based on sentiment analysis of social media data. The experimental results also checked the validation of the method using cross-validation based on the test results. The detailed validation calculations use three methods, namely accuracy, precision, and recall.

### A. Dataset and Labeling Process

This study uses data from YouTube comments by the crawling process. The data taken is based on the conditions of the COVID-19 pandemic regarding the free electricity program from the Indonesian government. The total dataset is 4,862 data of type integer. The data is divided into two categories, namely, positive and negative. Positive comments on this study have a total of 1651 positive comments from people who provide comments. Meanwhile, negative comments have a total data volume of 3211 negative comments from people who comment on the video. Positive comment data means that a person's condition has an anxiety value = 0 so that he has positive expectations. Meanwhile, negative comments mean someone has an anxious state. The classification results are shown in Figure 3.

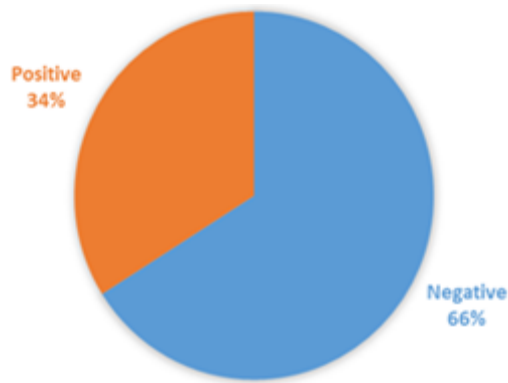

Figure 3. Percentage of initial labeling classification results in anxiety detection

Figure 3 illustrates the percentage of classification in detecting anxiety using the Indonesian sample based on YouTube comments. The sentences used are shown in Table 2. The dataset is classified into two categories, namely negative and positive. In Table 2, the sample data has a sample of five each for positive and negative comment data in Indonesian.

Tabel 2.
Sample YouTube comments with a sample of five for each negative and positive category

| No | Comment Text | Category |
|---|---|---|
| 1 | ini penyegelan banyak sekali tetangga sebelah. | Negative |
| 2 | nggak adil negara sama sama bayar pajak | Negative |
| 3 | Tak adil macam ini pln sama pengguna | Negative |
| 4 | Bagaimana dgn rmj dinas 450...tp listiknya byr sendiri tiap bln termsk PDAM | Negative |
| 5 | Setuju nih.. kebanyakan ngibul nasih gratisan aja.. | Negative |
| 6 | Ya gk pp di syukuri aj ya | Positive |
| 7 | Alhamdulillah saya bisa | Positive |
| 8 | Sabar nggeh, insyaallah rezekinya ada terus. Aamiin | Positive |
| 9 | Almhmdllh sdh dpat akses dan sdh dpat nmer tokennya... | Positive |

| 10 | Coba terus sampe keluar tokennya gan....saya semalam gitu...alhamdulillah sudah dapat | Positive |

## B. Experiment Results and Discussion

The experimental process was carried out using crawled data. The results of crawling data from YouTube comments are presented in Table 2 as a sample. The data is pre-processed to obtain clean data for further processing. This process fully uses the Python programming language. Classification result data was carried out using two data groupings, namely negative and positive. Python implementation uses a design as in the research in Figure 1 and Figure 2 for the process flow.

The process begins by reading the data and then preprocessing. Preprocessing aims to clean data. The resulting data is a core that has no noise. This process is carried out with several processes, including: tokenizing, filtering (slank word conversion, remove the number, remove stopword, remove the figure, remove duplicate), stemming, emoticon conversion. In the conversion process, emoticons have rules to improve data cleanliness. Additional rules besides the process are as follows:

a. There is a single figure word; it is necessary to add the word "support". The result is positive sentiment and emotional trust = 1

b. There is an exclamation mark, then the emotion of trust = 2

c. In one dominant capital letter sentence, angry emotions = 1

d. There is an angry emotion exclamation mark = 2

Here is a coding sample in Python to display the conversion of emotions from comment text. The conversion result is a complete sentence without emoticons. This is because the emoticon is converted into a complete sentence, what changes is the emoticon becomes the sentence ".... face_with_tears_of_joy. face_with_tears_of_joy. face_with_tears_of_joy." Because there are three emotions, they are all written down.

```
kalimat=(conv_emoticon("Subsidi listrik saya dicabut. R1 jadi R1M...Bpjs saya
 kena pengurangan.Auto kaya SAYA!! Dimata pak presiden pastinya😂😂😂"))
for kata in kalimat.split(): print(kata)
```

This process is a preprocessing, which is processed in full by coding using Python. In detail, Python commands and execution results can be seen in Figure 4 according to the design in Figure 1. The final result of this preprocessing is clean data. Result data is data that has meaning for the next process. The preprocessing eliminates meaningless/useless text. The concept of data cleaning can be seen from the column named "text" as the initial data and turned into the preprocessing data "clean_text" column. The clean data becomes the data used in anxiety detection so that the results obtained are better because vital information is used in the process.

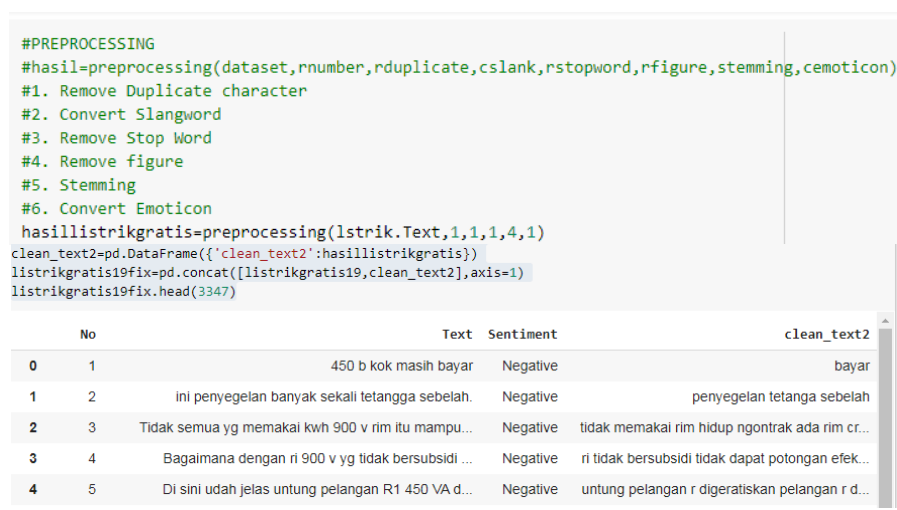

Figure 4. Preprocessing Results that have clean data

The preprocessing result data is in the form of clean data, which is processed by feature extraction. By the method presented in Figure 2, the methods used include Count-Vectorization and TF-IDF. This method's results are feature extraction data processed using machine learning methods, namely K-NN,

Bernoulli, Decision Tree Classifier, Support Vector Classifier, Random Forest, and XG-Boost. The code syntax in Python can be seen in Figure 5.

```
[ ] import xgboost as xgb
    #Feature Extraction
    cvec=CountVectorizer()
    tvec=TfidfVectorizer()

    #METHOD of MACHINE LEARNING
    clf1 = KNeighborsClassifier()
    clf2 = BernoulliNB()
    clf3 = tree.DecisionTreeClassifier()
    clf4 = SVC()
    clf5 = RandomForestClassifier()
    clf8 = xgb.XGBClassifier(objective="binary:logistic", random_state=42)
    methoddipakai=clf5
```

Figure 5. Use of feature extraction and machine learning methods in Python

```
model= Pipeline([('vectorizer',cvec),('classifier',methoddipakai)])
model.fit(x_train,y_train)
hasil = model.predict(x_test)
confusion_matrix(hasil,y_test, labels=["Positive", "Negative"])

array([[258,  80],
       [ 86, 549]])
```

```
accuracy_score(hasil,y_test)
0.829393627954779

precision_recall_fscore_support(hasil, y_test)
(array([0.87281399, 0.75      ]),
 array([0.86456693, 0.76331361]),
 array([0.86867089, 0.75659824]),
 array([635, 338]))
```

(a)        (b)

Figure 6. Examples of applying (a) Count-Vectorization feature extraction method with confusion matrix and (b) cross-validation calculations based on the Random Forest method

Based on Figure 5, the use of feature extraction methods and machine learning in anxiety detection is implemented in Figure 6. These implementation results are examples of applying the feature extraction method using Count-Vectorization, and the machine learning method using the random forest method. The validation process uses the application of formulas (1), (2), and (3), which are coded in Python with the functions provided in the python library.

Based on the matrix results in Figure 6, the results of the test results for each feature extraction method and machine learning are shown in Table 3. The highest accuracy result is the random forest method based on each feature extraction Count-Vectorization and TF-IDF with successive values are 84.99% and 82.63%. The highest scores for precision and recall are K-NN and XG-Boost, respectively. Although the precision or recall values for both are high, they cannot serve as a reference for anxiety detection recommendations. That is because the value of the two (recall/precision) is not balanced. Thus, the best consistency of all methods that have been processed with Python is the Random Forest (the result shown in Figure 7).

Apart from Random Forest, the application of machine learning methods with an accuracy of more than 80% and are at least balanced in calculating precision and recall can be considered. The methods that can be considered include Bernoulli, Decision Tree, and Support Vector Classifier. The three methods apart from having an accuracy of more than 80%, the precision and recall can be balanced. The precision and recall values were more than 60. Each result had differences that were not that far off, with a maximum difference of about 20% of the results.

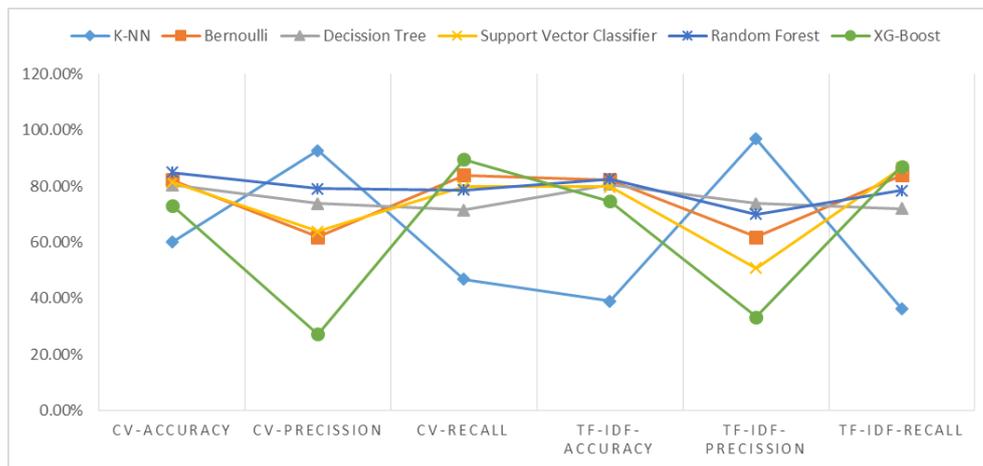

Figure 7. The graph of the Machine Learning method is based on cross-validation calculations

Tabel 3.
Results of the calculation of cross-validation from the Machine Learning Methods (such as K-NN, Bernoulli, Decision Tree, Support Vector Classifier, Random Forest and XG-Boost)

| Machine Learning | Feature Extraxtion | | | | | |
|---|---|---|---|---|---|---|
| | Count-Vectorization | | | TF-IDF | | |
| | Accuracy | Precission | Recall | Accuracy | Precission | Recall |
| K-NN | 60.12% | 92.73% | 46.77% | 39.05% | 96.80% | 36.39% |
| Bernoulli | 82.32% | 61.91% | 83.85% | 82.32% | 61.91% | 83.85% |
| Decission Tree | 80.36% | 73.83% | 71.54% | 80.57% | 73.83% | 71.95% |
| Support Vector Classifier | 81.50% | 63.95% | 79.71% | 79.85% | 50.87% | 86.63% |
| Random Forest | 84.99% | 79.06% | 78.61% | 82.63% | 70.05% | 78.50% |
| XG-Boost | 73.17% | 27.32% | 89.52% | 74.71% | 33.43% | 87.12% |

## V. CONCLUSION

The experiment and discussion results explain that the best machine learning method recommended in this study is the Random Forest method. Cross-Validation Testing, especially the level of accuracy, the random forest has an accuracy rate of close to 85% with Count-Vectorization feature extraction compared to all the machine learning methods used. The results showed that the sentiment data amounted to 3889 and 973 in testing and this study. The best precision test is K-NN, while the best recall is XG-Boost, but other values don't support it. Even so, the accuracy of applying machine learning methods with an accuracy of more than 80% includes Bernoulli, Decision Tree, Support Vector Classifier, and Random Forest. Thus, Random Forest can be the best reference for applying machine learning methods to detect anxiety based on social media data. In addition to the best accuracy, the random forest also has significant and bettef of precision and recall values are compared to other methods.